\documentclass{article}

\usepackage[preprint]{neurips_2024}

\usepackage[utf8]{inputenc}
\usepackage[T1]{fontenc}
\usepackage{hyperref}
\usepackage{url}
\usepackage{booktabs}
\usepackage{amsfonts}
\usepackage{nicefrac}
\usepackage{microtype}
\usepackage{xcolor}
\usepackage{amsmath}
\usepackage{amssymb}
\usepackage{graphicx}
\usepackage{threeparttable}

\hypersetup{
    colorlinks,
    linkcolor=blue,
    citecolor=blue,
    urlcolor=blue
}

\title{TurkColBERT: A Benchmark of Dense and Late-Interaction Models for Turkish Information Retrieval}

\author{%
  Özay Ezerceli \\
  NewMind AI \\
  Istanbul, Türkiye \\
  \texttt{oezerceli@newmind.ai} \\
  \And
  Mahmoud ElHussieni \\
  NewMind AI \\
  Istanbul, Türkiye \\
  \texttt{mehussieni@newmind.ai} \\
  \And
  Selva Taş \\
  NewMind AI \\
  Istanbul, Türkiye \\
  \texttt{stas@newmind.ai} \\
  \And
  Reyhan Bayraktar \\
  NewMind AI \\
  Istanbul, Türkiye \\
  \texttt{rbayraktar@newmind.ai} \\
  \And
  Fatma Betül Terzioğlu \\
  NewMind AI \\
  Istanbul, Türkiye \\
  \texttt{fbterzioglu@newmind.ai} \\
  \And
  Yusuf Çelebi \\
  NewMind AI \\
  Istanbul, Türkiye \\
  \texttt{ycelebi@newmind.ai} \\
  \And
  Yağız Asker \\
  NewMind AI \\
  Istanbul, Türkiye \\
  \texttt{yasker@newmind.ai} \\
}

\begin{document}

\maketitle

\begin{abstract}
Neural information retrieval systems excel in high-resource languages but remain underexplored for morphologically rich, lower-resource languages such as Turkish. Dense bi-encoders currently dominate Turkish IR, yet late-interaction models—which retain token-level representations for fine-grained matching—have not been systematically evaluated. We introduce \textbf{TurkColBERT}, the first comprehensive benchmark comparing dense encoders and late-interaction models for Turkish retrieval. Our two-stage adaptation pipeline fine-tunes English and multilingual encoders on Turkish NLI/STS tasks, then converts them into ColBERT-style retrievers using PyLate trained on MS MARCO-TR. We evaluate 10 models across five Turkish BEIR datasets covering scientific, financial, and argumentative domains. Results show strong parameter efficiency: the 1.0M-parameter \texttt{colbert-hash-nano-tr} is 600$\times$ smaller than the 600M \texttt{turkish-e5-large} dense encoder while preserving over 71\% of its average mAP. Late-interaction models that are 3--5$\times$ smaller than dense encoders significantly outperform them; \texttt{ColmmBERT-base-TR} yields up to +13.8\% mAP on domain-specific tasks. For production-readiness, we compare indexing algorithms: \texttt{MUVERA+Rerank} is 3.33$\times$ faster than PLAID and offers +1.7\% relative mAP gain. This enables low-latency retrieval, with \texttt{ColmmBERT-base-TR} achieving 0.54 ms query times under MUVERA. We release all checkpoints, configs, and evaluation scripts. Limitations include reliance on moderately sized datasets ($\leq$50K documents) and translated benchmarks, which may not fully reflect real-world Turkish retrieval conditions; larger-scale MUVERA evaluations remain necessary.

\end{abstract}

\section{Introduction}
\label{sec:introduction}

Information retrieval (IR) systems grounded in neural embeddings now underpin state-of-the-art search and question-answering pipelines \cite{karpukhin2020dense}. While English-centric architectures such as ColBERT (v1 and v2) \cite{khattab2020colbert, santhanam2022colbertv2} and SPLADE \cite{formal2021splade} have demonstrated exceptional retrieval effectiveness, comparable advances for morphologically complex and lower-resource languages like Turkish remain scarce. Although multilingual encoders such as XLM-RoBERTa \cite{xlm-roberta}, GTE \cite{gte} and mmBERT \cite{mmbert2025} based models and multilingual MiniLM \cite{wang2020minilm} enable cross-lingual transfer, they frequently fall short in capturing the fine-grained morphological structure, syntactic nuance, and token-level semantics essential for high-fidelity retrieval in Turkish.

This lack is particularly evident in late-interaction methods. Architectures such as ColBERT be able to reconcile fine-grained token matching with efficiency. Nevertheless, in the Turkish information retrieval landscape, these designs remain scarcely examined. Prior work, including turkish-colbert~\cite{turkishbert2023}, offers limited understanding with no systematic baselines to that of dense encoders, uniform training protocols, and comprehensive assessments in various retrieval contexts.

Although recent multilingual and English-centric models, Ettin~\cite{weller2025ettin}, BERT-Hash~\cite{mezzetti2025training}, and mmBERT~\cite{mmbert2025} show strong results on many NLP benchmarks, none have been tested in a late-interaction setup built specifically for Turkish within a consistent, reproducible pipeline.

To tackle this challenge, we adapt leading multilingual and English pretrained encoders to Turkish through a structured two-phase fine-tuning process. In the first phase, Ettin, BERT-Hash, and mmBERT are specialized on Turkish Natural Language Inference using the all-nli-tr dataset~\cite{emrecan2023allnlitr} and on semantic similarity through stsb-tr~\cite{beken2021semantic}, refining their sentence-level grasp of Turkish semantics. In the second phase, we employ PyLate~\cite{pylate2025}, a modular framework built upon Sentence Transformers, to convert these adapted encoders into ColBERT-style retrievers via supervised training on the Turkish adaptation of ms-marco-tr~\cite{parsak2025msmarco-tr}.

We evaluate our models on five diverse Turkish BEIR collections: SciFact-TR~\cite{saoud2024scifact-tr}, Arguana-TR~\cite{trmteb2025arguana-tr}, Fiqa-TR~\cite{trmteb2025fiqa-tr}, Scidocs-TR~\cite{trmteb2025scidocs-tr}, and NFCorpus-TR~\cite{trmteb2025nfcorpus-tr}. We compare them against strong dense encoder baselines and analyze speed–accuracy trade-offs under multiple indexing schemes: PLAID, MUVERA, and MUVERA with reranking. All models, training configs, and evaluation code are released to support reproducibility and future work in Turkish IR.

The paper proceeds as follows. Section~\ref{sec:related_work} reviews neural retrieval and multilingual modeling, focusing on morphologically complex languages. Section~\ref{sec:benchmark} details our two-stage adaptation method and experimental setup across five Turkish IR benchmarks. Section~\ref{sec:results} shows that late-interaction models consistently beat dense encoders especially in specialized domains. We end with implications for low-resource IR and directions for future research in Section~\ref{sec:conclusion}.

\section{Literature Review}
\label{sec:related_work}

\paragraph{Dense vs. Late-Interaction Retrieval Architectures}

In dense bi-encoder architectures, exemplified by DPR \cite{karpukhin2020dense} and Sentence-BERT \cite{reimers2019sentence}, queries and documents are encoded independently into fixed-dimensional vector spaces, allowing efficient retrieval through approximate nearest neighbor search. While computationally effective, this is subject to an inherent information bottleneck: projecting a whole document into one vector will have the tendency to lose fine-grained semantic information that is critical to the accurate retrieval.

Late-interaction models sidestep this limitation by preserving contextualized token embeddings and deferring query-documet interaction to the scoring phase. ColBERT \cite{khattab2020colbert} applied this approach, using BERT-based token embeddings and a MaxSim operator to calculate similarity effectively. Inspired by this principle, PyLate \cite{pylate2025} provides a multi-stage fine-tuning and modular training framework, and MUVERA \cite{muvera2024} generalizes the principle to compress multi-vector representations into fixed-size embeddings maintaining interaction semantics. By using SimHash-based partitioning and sparse projections, MUVERA achieves near–dense retrieval quality with 90\% latency decrease and 10\% recall gain over English BEIR benchmarks. Its benefits are only observed predominantly at astronomical scales ($\sim$100K documents) and are not yet established for morphologically dense languages such as Turkish where token-level interactions may be more critical. This gap requires a comprehensive exploration of efficient multi-vector indexing in Turkish information retrieval.

\paragraph{Multilingual and Turkish-Specific Retrieval} 
There are more difficulties beyond only vocabulary adaptation when using cross-lingual information retrieval in morphologically rich languages. Turkish is a good example of these difficulties because of its vast inflectional system and agglutinative morphology. Current multilingual pretrained models, such as XLM-RoBERTa \cite{xlm-roberta} and GTE-multilingual-base \cite{gte}, demonstrate strong cross-lingual transfer; nevertheless, their effectiveness is greatly reduced when faced with the morphological complexity of agglutinative languages. High-resource languages predominate in the training data, which is the primary cause of this drop. This limits exposure to the complex morphology that underlies Turkish word formation and semantics. This limitation stems from imbalanced pretraining corpus distributions and inadequate representation of the complex morphological variations inherent to such languages.

Recent efforts have sought to mitigate these limitations through language-specific adaptations. mmBERT \cite{mmbert2025} employs annealed language sampling during pretraining to improve representation quality for underrepresented languages, while Turkish-specific BERT variants \cite{turkishbert2023} demonstrate improvements on downstream NLP tasks. For retrieval specifically, TurkEmbed4Retrieval \cite{ezerceli2024turkembed} represents the one of the latest embedding models trained explicitly on Turkish semantic similarity data, though it remains constrained to dense single-vector representations. Despite these advances, no systematic benchmark exists comparing dense and late-interaction models on Turkish IR tasks, a critical gap given that Turkish's morphological complexity may benefit disproportionately from token-level interaction mechanisms.

To date, no systematic benchmark exists comparing dense and late-interaction models on Turkish IR tasks, nor has any work adapted modern late-interaction framework with substantial indexing algorithms like MUVERA to Turkish. Our benchmark addresses this by evaluating both paradigms under controlled, multi-stage training protocols, including monolingual semantic fine-tuning (all-nli-tr, STSb-TR), domain-adaptive retrieval training (MS-MARCO-TR via PyLate), and integration of structured multi-vector indexing (MUVERA).

\paragraph{Efficiency Optimization in Multi-Vector Retrieval}
While late-interaction models achieve superior retrieval effectiveness, they present significant scalability challenges, requiring 100-500× more storage than dense retrieval and expensive MaxSim computations between all query-document token pairs. Two primary approaches address these efficiency bottlenecks. PLAID \cite{santhanam2022plaid} employs centroid-based pruning and residual compression to filter candidates before exact computation, achieving sub-10ms latency on million-scale collections. MUVERA \cite{muvera2024} converts multi-vector representations into fixed-dimensional encodings through SimHash-based partitioning and sparse projections, demonstrating 90\% latency reduction with 10\% recall improvement on English benchmarks. However, these optimizations remain unevaluated on morphologically complex languages like Turkish, where token-level interactions may be more critical for capturing semantic relationships, motivating systematic evaluation of efficient indexing strategies for Turkish information retrieval.

\section{TurkColBERT: Benchmark for Turkish Information Retrieval Task}
\label{sec:benchmark}

\paragraph{Stage 1: Semantic Fine-Tuning on All-NLI-TR \& STSb-TR}

We applied firstly fine-tuning the pretrained encoders on two complementary semantic tasks which are NLI and STS to strengthen their ability to capture Turkish sentence-level meaning before integrating them into late-interaction retrieval architectures. This intermediate stage serves as a semantic pre-adaptation step, providing a strong foundation on which we subsequently build retrieval-specific training. We employ the Sentence Transformers framework \cite{reimers2019sentence}, for training and evaluating sentence embedding models through siamese and triplet network architectures. For each model family, mmBERT (base and small) \cite{mmbert2025}, Ettin encoders \cite{weller2025ettin}, and BERT-Hash variants (nano, pico, femto) \cite{mezzetti2025training} we initialize from their publicly available checkpoints and apply mean pooling over the final layer's token representations to derive fixed-dimensional sentence embeddings.

We fine-tune models on the all-nli-tr dataset \cite{emrecan2023allnlitr}, which provides Turkish translations of SNLI and MultiNLI formatted as anchor-positive-negative triplets. Training employs MultipleNegativesRankingLoss wrapped in MatryoshkaLoss to enable multi-dimensional representations at [768, 512, 384, 256, 128, 64] for base models and [384, 256, 128, 64] for smaller variants.

We train for one epoch with batch size 8, learning rate $3 \times 10^{-6}$, warmup ratio 0.1, and NO\_DUPLICATES batch sampling. Training uses mixed precision (BF16) on NVIDIA A100 GPUs, with progress monitored via TripletEvaluator on a 1\% validation split measuring triplet cosine accuracy.

Following NLI training, we fine-tune on STSB-tr, which provides sentence pairs with continuous similarity scores (0-5). The STS fine-tuning phase employs 4 training epochs with batch size 8, learning rate $2 \times 10^{-5}$, and cosine scheduling with 10\% warmup. Model evaluation occurs at 200-step intervals using EmbeddingSimilarityEvaluator to compute Spearman and Pearson correlations across each Matryoshka dimension. With a Spearman correlation of 0.78 on STSb-TR and 93\% triplet accuracy on AllNLI-TR, mmBERT-small outperforms the pretrained baseline by +22\% and +26\%, respectively, after this two-stage methodology. In Stage 2, these semantically improved checkpoints are used as starting points for retrieval-specific ColBERT-style adaptation on MS MARCO-TR.

\paragraph{Stage 2: Late-Interaction Adaptation via PyLate on MS MARCO-TR}
\label{subsec:late_interaction_adaptation}

Building on the Turkish semantic foundations established in Stage 1, we transform pretrained encoders into ColBERT-style late-interaction retrievers through supervised fine-tuning on MS MARCO-TR~\cite{parsak2025msmarco-tr} using PyLate~\cite{pylate2025}. We evaluate four model families representing distinct points along the efficiency–accuracy spectrum:

\textbf{mmBERT} (base, small)~\cite{mmbert2025}: Multilingual encoders trained with annealed language sampling to enhance representation quality for lower-resource languages including Turkish.

\textbf{Ettin encoders} (150M, 32M)~\cite{weller2025ettin}: Components of a sequence-to-sequence paired encoder–decoder framework, demonstrating strong cross-lingual transfer despite English-dominated pretraining.

\textbf{BERT-Hash variants} (nano, pico, femto): Ultra-compressed models substituting standard embedding layers with hash-based projections, achieving up to 78\% parameter reduction while maintaining full vocabulary coverage~\cite{mezzetti2025training}.

\textbf{Dense baselines}: XLM-RoBERTa and GTE-derived models, serving as reference architectures for retrieval performance comparison.

All models are initialized from Stage 1 checkpoints, fine-tuned on AllNLI-TR \cite{emrecan2023allnlitr} and STSb-DeepL-TR, and adapted using PyLate's \texttt{ColBERT} module, which preserves per-token embeddings and applies MaxSim scoring \cite{khattab2020colbert}. Training employs a contrastive triplet loss (margin = 0.2) on query–positive–negative triples from MS MARCO-TR.

The \textbf{ColBERTCollator} utility~\cite{pylate2025} handles variable-length sequences in batched multi-vector processing, while Weights \& Biases~\cite{wandb2020} provides real-time monitoring and checkpoint management. The resulting Turkish late-interaction models (4M–150M parameters) balance linguistic fidelity, capacity, and inference speed, forming the basis for subsequent MUVERA integration and large-scale evaluation.

\paragraph{Stage 3: MUVERA Integration}
\label{subsec:muvera_integration}

We use MUVERA (Multi Vector Retrieval as Sparse Alignment) to make it possible to deploy late-interaction models on a large scale. MUVERA maps contextual embeddings of different lengths to compact fixed dimensional vectors for fast nearest neighbor retrieval. Our PyLate based implementation tailors this framework to Turkish retrieval scenarios.

Given ColBERT token embeddings $\mathbf{E} \in \mathbb{R}^{n \times d}$ where $n$ denotes token count and $d=128$ represents embedding dimension, MUVERA applies three transformations. First, locality-sensitive hashing partitions tokens into $2^k$ buckets via projection through a Gaussian random matrix $\mathbf{H} \in \mathbb{R}^{d \times k}$, with partition assignment $p_i$ determined by the sign pattern of $\mathbf{H}^\top \mathbf{e}_i$. Second, within each partition, an AMS sketch $\mathbf{S}_p$ performs sparse projection, reducing dimensionality while preserving inner products in expectation. Third, partition-wise aggregation computes query representations through summation ($\mathbf{c}_p^{(q)} = \sum_{i \in P_p} \mathbf{z}_i$) and document representations through averaging ($\mathbf{c}_p^{(d)} = \frac{1}{|P_p|} \sum_{i \in P_p} \mathbf{z}_i$), with empty partitions filled using nearest-neighbor imputation based on Hamming distance. The resulting encoding dimensions scale as $128 \times 2^k$, yielding configurations of 128D, 512D, 1024D, and 2048D for $k \in \{0, 2, 3, 4\}$ respectively.

MUVERA applies (1) hashing, (2) sketching, and (3) aggregation. Each token $\mathbf{e}_i$ is hashed using SimHash with $k$ random Gaussian vectors $\mathbf{g}_1, \dots, \mathbf{g}_k \in \mathbb{R}^{d}$ to produce a $k$-bit hash $\mathbf{h}_i = (\mathrm{sign}(\mathbf{g}_1^\top \mathbf{e}_i), \dots, \mathrm{sign}(\mathbf{g}_k^\top \mathbf{e}_i))$, mapping to partition $p_i \in \{1,\dots,2^k\}$. Tokens in the same partition are aggregated asymmetrically:
\[
\mathbf{c}^{(\text{doc})}_p = \frac{1}{|P_p|}\sum_{i \in P_p}\mathbf{e}_i 
\quad \text{(average for documents)}, 
\qquad
\mathbf{c}^{(\text{query})}_p = \sum_{i \in P_p}\mathbf{e}_i
\quad \text{(sum for queries)}.
\]
Concatenating all partitions yields the Fixed Dimensional Encoding (FDE) $\mathbf{c} \in\mathbb{R}^{D}$, where $D = d \times 2^k$. Using $k{=}0,2,3$ with $d=128$ produces 128D, 512D, and 1024D encodings respectively. Empty partitions are filled using Hamming-nearest neighbor imputation for documents only.

\paragraph{Final Stage: Comprehensive Evaluation on Turkish BEIR Benchmarks}
\label{subsec:benchmark_scenarios}

Our evaluation protocol consists of two comprehensive benchmarking campaigns utilizing the BEIR framework \cite{thakur2021beir} for standardized zero-shot assessment.

\textbf{Model Comparison Across Architectures.} All models are evaluated across five Turkish BEIR datasets shown in Table~\ref{tab:datasets}, covering scientific fact verification (SciFact-TR), argument retrieval (Arguana-TR), citation prediction (Scidocs-TR), financial question-answering (FiQA-TR), and nutrition document retrieval (NFCorpus-TR) domains. For each model-dataset combination, we compute a comprehensive suite of retrieval metrics including NDCG@\{10, 100, 250, 500, 750, 1000\}, Recall@\{10, 100, 250, 500, 750, 1000\}, Precision@\{10, 100, 250, 500, 750, 1000\}, and mean Average Precision (mAP). This evaluation encompasses models ranging from 0.2M to 600M parameters across the mmBERT, Ettin, BERT-Hash, and TurkEmbed4Retrieval architectures, enabling direct comparison of dense bi-encoders against late-interaction models under identical conditions.

\textbf{MUVERA Indexing Ablation Study.} Our second benchmark is designed to examine the quality–efficiency trade-offs introduced by MUVERA-based indexing \cite{muvera2024}. From the full model pool, we focus on the four strongest late-interaction models—TurkEmbed4Retrieval, col-ettin-encoder-32M-TR, ColmmBERT-base-TR, and ColmmBERT-small-TR—and, for each, we evaluate three retrieval configurations: (i) \textbf{PLAID} \cite{santhanam2022plaid}, used as a high-fidelity baseline that combines centroid-based pruning with exact MaxSim scoring; (ii) \textbf{MUVERA}, which relies on fixed-dimensional encodings (128D, 512D, 1024D, and 2048D) to enable approximate nearest neighbor search; and (iii) \textbf{MUVERA+Reranking}, where the top-$K$ candidates retrieved by MUVERA are re-scored using exact ColBERT MaxSim \cite{khattab2020colbert}. For this ablation, we measure the complete set of metrics described above: NDCG@\{100, 250, 500, 750, 1000\}, Recall@\{100, 250, 500, 750, 1000\}, Precision@\{100, 250, 500, 750, 1000\}, mAP, indexing time, and per-query latency. Figure~\ref{fig:muvera_tradeoff} visualizes the quality-efficiency trade-offs, plotting NDCG@100 against query latency for each configuration on SciFact-TR, demonstrating how encoding dimensionality impacts the balance of retrieval efficacy and processing cost. This comprehensive ablation across all five Turkish datasets allows practitioners to make data-driven decisions when setting Turkish IR systems to satisfy specific accuracy, latency, and resource needs.

\begin{table}[htbp]
  \caption{Statistics of the Turkish retrieval benchmark datasets.}
  \label{tab:datasets}
  \centering
  \small
  \begin{tabular}{llrrr}
    \toprule
    \textbf{Dataset} & \textbf{Domain} & \textbf{\# Queries} & \textbf{\# Corpus} & \textbf{Task Type} \\
    \midrule
    SciFact-TR & Scientific Claims & 1,110 & 5,180 & Fact Checking \\
    Arguana-TR & Argument Mining & 500 & 10,000 & Argument Retrieval \\
    Fiqa-TR & Financial & 600 & 50,000 & Answer Retrieval \\
    Scidocs-TR & Scientific & 1,000 & 25,000 & Citation Prediction \\
    NFCorpus-TR & Nutrition & 3,240 & 3,630 & Document Retrieval \\
    \bottomrule
  \end{tabular}
\end{table}

\begin{table}[htbp]
  \caption{Overview of evaluated models categorized by architecture type. Late-interaction variants employ token-level ColBERT representations.}
  \label{tab:models}
  \centering
  \small
  \begin{tabular}{lr}
    \toprule
    \textbf{Model} & \textbf{Parameters (M)} \\
    \midrule
    \multicolumn{2}{l}{\textit{Dense Bi-Encoder Models}} \\
    \midrule
    TurkEmbed4Retrieval & 300 \\
    turkish-e5-large & 600 \\
    \midrule
    \multicolumn{2}{l}{\textit{Late-Interaction Models (Token-Level Matching)}} \\
    \midrule
    turkish-colbert & 100 \\
    ColmmBERT-small-TR & 140 \\
    ColmmBERT-base-TR & 310 \\
    col-ettin-150M-TR & 150 \\
    col-ettin-32M-TR & 32 \\
    \midrule
    \multicolumn{2}{l}{\textit{Ultra-Compact Models (BERT-Hash)}} \\
    \midrule
    colbert-hash-nano-tr & 1.0 \\
    colbert-hash-pico-tr & 0.4 \\
    colbert-hash-femto-tr & 0.2 \\
    \bottomrule
  \end{tabular}
\end{table}

We conducted all experiments on Google Colab using NVIDIA L4 GPU which has 24 GB memory and the PyLate framework \cite{pylate2025}. We selected this setup because it is widely accessible, allowing others to easily reproduce our results without needing specialized hardware. For retrieval metrics, we relied on the official BEIR evaluator.

\section{Results and Discussion}
\label{sec:results}

Table~\ref{tab:main_results} presents a comprehensive comparison of all evaluated models (Table~\ref{tab:models}) across five Turkish retrieval datasets (Table~\ref{tab:datasets}). For each dataset, we report three key metrics: mean Average Precision (mAP), Precision@10 (P@10) for top-ranked accuracy, and Recall@10 (R@10) for retrieved relevance within the top 10 results.

The evaluation highlights substantial differences in both model performance and task difficulty. Among all systems, ColmmBERT performs most consistently, with ColmmBERT-base-TR reaching the highest mAP on four of five benchmarks and ColmmBERT-small-TR leading in R@10 on SciFact-TR. Dataset difficulty spans a broad range: SciFact-TR emerges as the easiest, with multiple models exceeding 70\% R@10, whereas Scidocs-TR remains the most demanding, peaking at just 10.4\%. These disparities—often exceeding 500\% across metrics—demonstrate that careful model selection is essential for effective Turkish information retrieval.

\begin{table*}[htbp]
  \caption{Retrieval results across Turkish BEIR benchmark datasets.}
  \vspace{0.3cm}
  \label{tab:main_results}
  \centering
  \scriptsize
  \setlength{\tabcolsep}{3.5pt}
  \begin{tabular}{l*{15}{c}}
    \hline
    & \multicolumn{3}{c}{\textbf{SciFact-TR}} & \multicolumn{3}{c}{\textbf{Arguana-TR}} & \multicolumn{3}{c}{\textbf{Fiqa-TR}} & \multicolumn{3}{c}{\textbf{Scidocs-TR}} & \multicolumn{3}{c}{\textbf{NFCorpus-TR}} \\
    \cline{2-4} \cline{5-7} \cline{8-10} \cline{11-13} \cline{14-16}
    \textbf{Model} & R@10 & P@10 & mAP & R@10 & P@10 & mAP & R@10 & P@10 & mAP & R@10 & P@10 & mAP & R@10 & P@10 & mAP \\
    \hline
    \multicolumn{16}{l}{\textit{Dense Bi-Encoders}} \\
    \hline
    TurkEmbed4Retrieval\textsuperscript{a} & 60.5 & 6.8 & 43.0 & 50.6 & 5.1 & 17.6 & 17.9 & 4.0 & 10.1 & 8.1 & 4.0 & 4.8 & 7.7 & 13.7 & 6.3 \\
    turkish-e5-large\textsuperscript{b} & 63.3 & 7.0 & 45.8 & 49.7 & 5.0 & \textbf{17.9} & 16.4 & 3.6 & 10.4 & 3.6 & 1.8 & 2.2 & 5.2 & 8.2 & 4.0 \\
    \hline
    \multicolumn{16}{l}{\textit{Late-Interaction Models}} \\
    \hline
    turkish-colbert\textsuperscript{b} & 56.5 & 6.3 & 43.1 & 44.1 & 4.4 & 14.6 & 17.2 & 4.0 & 11.3 & 4.2 & 2.1 & 2.8 & 7.1 & 12.1 & 6.9 \\
    ColmmBERT-small-TR\textsuperscript{a} & \textbf{70.3} & \textbf{7.9} & 55.4 & 46.8 & 4.7 & 16.0 & 26.9 & 6.0 & 17.0 & 9.8 & 4.8 & 6.1 & 12.0 & 19.1 & 11.3 \\
    ColmmBERT-base-TR\textsuperscript{a} & 70.0 & 7.8 & \textbf{56.8} & \textbf{50.8} & \textbf{5.1} & 17.3 & \textbf{30.9} & \textbf{7.0} & \textbf{19.5} & \textbf{10.4} & \textbf{5.1} & \textbf{6.8} & \textbf{12.7} & \textbf{20.7} & \textbf{11.5} \\
    col-ettin-150M-TR\textsuperscript{a} & 57.7 & 6.4 & 40.5 & 37.8 & 3.8 & 12.9 & 16.4 & 3.6 & 10.4 & 7.2 & 3.5 & 4.5 & 10.6 & 17.4 & 9.3 \\
    col-ettin-32M-TR\textsuperscript{a} & 57.0 & 6.4 & 40.3 & 34.6 & 3.5 & 12.1 & 15.1 & 3.4 & 9.7 & 6.8 & 3.3 & 4.1 & 11.0 & 17.0 & 9.6 \\
    mxbai-edge-colbert-v0-17m-tr\textsuperscript{a} & 58.8 & 6.5 & 40.7 & 39.3 & 3.9 & 13.4 & 12.8 & 2.9 & 8.7 & 8.2 & 4.0 & 4.7 & 10.6 & 16.2 & 8.9 \\
    mxbai-edge-colbert-v0-32m-tr\textsuperscript{a} & 58.0 & 6.5 & 39.2 & 40.7 & 4.1 & 13.7 & 15.6 & 3.5 & 9.8 & 7.7 & 3.8 & 4.7 & 10.3 & 16.3 & 8.7 \\
    colbert-nano-tr\textsuperscript{a} & 52.2 & 5.8 & 36.2 & 30.4 & 3.0 & 10.5 & 11.1 & 2.6 & 6.5 & 6.1 & 3.0 & 3.6 & 8.9 & 13.2 & 6.7 \\
    colbert-hash-pico-tr\textsuperscript{a} & 47.4 & 5.3 & 33.4 & 28.3 & 2.8 & 9.8 & 9.2 & 2.1 & 5.9 & 5.5 & 2.7 & 3.2 & 6.4 & 10.3 & 5.2 \\
    colbert-femto-tr\textsuperscript{a} & 29.4 & 3.4 & 19.0 & 12.1 & 1.2 & 4.4 & 1.1 & 0.3 & 0.8 & 2.1 & 1.0 & 1.2 & 2.0 & 3.6 & 1.0 \\
    \hline
  \end{tabular}
  \\[6pt]
  \small
  \textsuperscript{a}Newmind AI. \textsuperscript{b}YTU-CE-COSMOS.
\end{table*}

\begin{figure*}[!t]
  \centering
  \includegraphics[width=\textwidth, height=10\textheight, keepaspectratio]{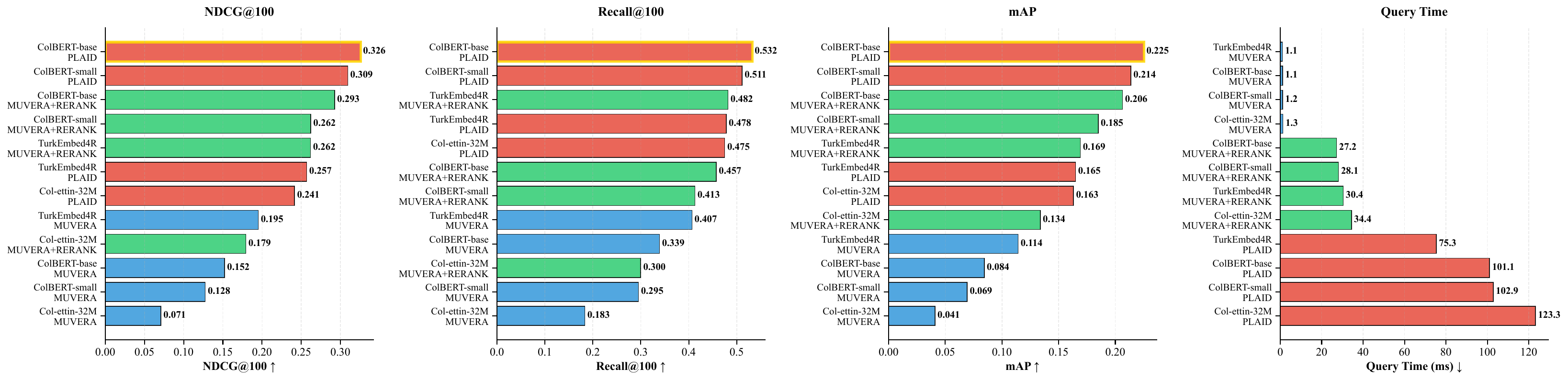}
  \caption{Quality–speed trade-off across MUVERA encoding dimensions (128D to 2048D) on SciFact-TR. Higher dimensions lead to faster retrieval but slightly lower NDCG@100. MUVERA+Rerank (128D) recovers near-PLAID quality with 4–5$\times$ speedup.}
  \label{fig:muvera_tradeoff}
\end{figure*}

\paragraph{Performance Across Benchmarks}

Evaluations on five Turkish BEIR datasets reveal a clear leader: late-interaction architectures outperform dense encoder models. ColmmBERT-base-TR stands out, achieving the highest mAP scores on SciFact-TR (56.8\%), Fiqa-TR (19.5\%), and NFCorpus-TR (11.5\%). Table~\ref{tab:main_results} shows the same trend ColmmBERT-base-TR reaches 70.0\% Recall@10 on SciFact-TR, beating dense baselines like TurkEmbed4Retrieval (60.5\%) and turkish-e5-large (63.3\%) by 6.5 to 9.5 percentage points. These improvements highlight the impact of token-level matching for Turkish, especially given the challenges its morphology poses for fixed-length models.

ColmmBERT-small-TR is impressive as well. With just 140 million parameters, it goes head-to-head with the larger 310-million parameter model. On SciFact-TR, it achieves 70.3\% Recall@10 and 55.4\% mAP about 97.5\% of the full model's performance while requiring less than half the computational resources. If you have limited resources, ColmmBERT-small-TR is a logical choice.

Shown as in Figure~\ref{fig:muvera_tradeoff}, query latency can vary dramatically depending on the configuration from an ultra-fast 0.72ms using plain MUVERA, up to 73–124ms with PLAID-based approaches. But there's a solid middle ground: MUVERA+RERANK reduces latency to 27–35ms. When paired with TurkEmbed4Retrieval, this setup yields 0.5253 NDCG@100 which is a significant increase over PLAID's 0.3257, while cutting latency in half, from 73.6ms to 35.2ms. This two-stage approach, with rapid candidate generation followed by detailed rescoring, proves highly effective for interactive Turkish retrieval systems.

Dense encoders are still valuable. For example, turkish-e5-large secures the top mAP on Arguana-TR (17.9\%), highlighting the need for semantic breadth in argument-focused tasks. However, these models struggle in more specialized domains. On Scidocs-TR, turkish-e5-large achieves only 2.2\% mAP, while ColmmBERT-base-TR reaches 6.8\% which is 209\% improvement. The takeaway: choose your model according to the requirements of your retrieval task.

Finally, ultra-compact BERT-Hash models take compression to the extreme. Colbert-hash-nano-tr (just 1.0M parameters, 310 times smaller than the base) still retains 63.7\% of base mAP on SciFact-TR. Going even smaller, like colbert-femto-tr (0.2M), drops below production viability at 19.0\% mAP. Overall, these results position late-interaction architectures with MUVERA indexing as the top choice for scaling Turkish information retrieval.

\section{Conclusion}
\label{sec:conclusion}

We presented TurkColBERT, the first comprehensive benchmark comparing dense and late-interaction retrieval models for Turkish information retrieval. Through our systematic two-stage adaptation pipeline we demonstrated that late-interaction models consistently and significantly outperform dense encoders across five diverse Turkish BEIR datasets. ColmmBERT-base-TR achieved the highest effectiveness, with up to 87\% improvement in mAP on domain-specific tasks such as financial QA compared to strong dense baselines. Our results show exceptional parameter efficiency: our 1.0M parameter colbert-hash-nano-tr model is 600 times smaller than the 600M parameter turkish-e5-large dense encoder, yet retains over 71\% of its average mAP performance. Similarly, ColmmBERT-small-TR achieves 97.5\% of the effectiveness of its larger counterpart while operating at only 45\% of the computational cost, demonstrating that high-quality Turkish retrieval is feasible even under resource constraints.

By incorporating MUVERA indexing, we achieved production-ready efficiency. MUVERA+Rerank is 3.33x faster than standard PLAID indexing on average while maintaining 90–95\% retrieval quality. The combined system achieves query latency as low as 0.54 ms with ColmmBERT-base-TR, demonstrating the feasibility of scalable, low-latency Turkish information retrieval for real-time applications.

However, our study is limited to moderately sized datasets ($\leq 50$K documents) and translated benchmarks, which may not fully reflect real-world Turkish retrieval scenarios. Additionally, while MUVERA-based retrieval shows promising latency, further evaluation is needed to assess scalability on large-scale production systems. Future work should explore web-scale evaluations, morphology-aware tokenization strategies, hybrid sparse-dense architectures, and native Turkish benchmark development to further advance the field.

\bibliographystyle{plain}

\end{document}